\documentclass[runningheads,a4paper]{llncs}

\usepackage{amssymb}
\usepackage{graphicx}
\usepackage{tikz}
\usetikzlibrary{arrows.meta,calc,fit,positioning}
\usepackage{hyperref}
\usepackage[margin=1.1in]{geometry}
\hypersetup{hidelinks}
\newcommand{\keywords}[1]{\par\addvspace\baselineskip
\noindent\keywordname\enspace\ignorespaces#1}

\newcommand{\supplementaryurl}{https://doi.org/10.5281/zenodo.21310200}

\begin{document}

\mainmatter

\title{One Timeline, Many Renderings:\\ A Wolfram Language Paclet\\ for heterogeneous musical output}

\titlerunning{One Timeline, Many Renderings}

\author{Francesco Vitucci\textsuperscript{1} \and Michele Lorusso\textsuperscript{2} \and Francesco Scagliola\textsuperscript{3}}
\authorrunning{F. Vitucci et al.}
\institute{\textsuperscript{1,2,3}Conservatorio di Musica ``N. Piccinni'' di Bari\\
\textsuperscript{1}\email{francescovitucci1@gmail.com}\\
\textsuperscript{2}\email{michelelorusso99@outlook.com}\\
\textsuperscript{3}\email{francesco.scagliola@gmail.com}}

\maketitle

\begin{abstract}
One algorithmic composition may require a Csound score, engraved notation,
real-time control, and a rehearsal click. Authored separately, their timelines
drift. Temporal System is a Wolfram Language paclet that instead compiles one
immutable store of typed entities on a rational beat timeline through
backend-specific contracts. It emits Csound synthesis, beta MusicXML~4.0, OSC
control, and click artifacts that remain synchronized because they share that
store. Conversion to seconds, samples, or hertz occurs only at render time.
Csound notes use stable named instruments in external \texttt{.orc} files;
curves become k-rate signals declared against score p-fields. The click backend
derives rehearsal audio from the same meter and tempo and reuses the Csound
serializer. We describe the temporal, semantic, and rendering-contract layers,
their practical trade-offs, and the limits of this proprietary authoring
environment within an otherwise open-source ecosystem.
The archived supplement exposes the reported outputs pending paclet release.
\end{abstract}

\section{Introduction}

Computing composers long ago separated structure from sound. Tenney generated
note lists for the \textsc{music~n} programs \cite{tenney,mathews}; Koenig's
\emph{Project~1/2} computed tables realized later by players or the studio
\cite{koenig}. In both, a program owns time and structure while another system
owns sound---the orchestra--score split Csound institutionalizes
\cite{csoundbook}.

Today a mixed work may need \texttt{.csd}, MusicXML, timed live control, and a
click. Their seconds, divisions, milliseconds, and samples diverge when edited
separately. Computer-assisted composition environments address parts of this
\cite{openmusic,taube}; Temporal System, a Wolfram Language paclet
\cite{wolfram}, makes one timeline authoritative and every output derived.

Multiple output and exact time are not new: \emph{music21} has exact fractional
offsets and exports MusicXML, MIDI, Lilypond, and text \cite{music21}; Abjad
drives Lilypond \cite{abjad}; MEI/Verovio project a canonical score
\cite{mei,verovio}; \emph{bach} realizes timelines in Max \cite{bach}; and
Tidal uses rational cycles for live scheduling \cite{tidal}. Temporal System's
narrower contribution is their combination across explicit contracts for
notation, offline Csound, OSC, and click output, deferring unit conversion to
each serializer. Wolfram Language~15.0 was chosen because paclet packaging,
symbolic transformations, notebooks, and timeline graphics coexist where the
authors already develop the music. More fundamentally, Mathematica is an ideal
instrument for investigating Stephen Wolfram's computational vision, including
the rule-based and multi-computational perspectives developed in his accounts
of ruliology and multicomputation \cite{wolframruliology,wolframmulticomputation}.
This reduces integration boundaries; exact
arithmetic itself is not exclusive, since Python's \texttt{Fraction} and
Julia's \texttt{Rational} support the same model.

Its strict decisions are: exact Rational onsets in an immutable typed store;
authored meter; session tuning; external \texttt{.orc} instruments; and a
permissive shared store constrained by each renderer's contract.

The choice still has an accessibility and sustainability cost: authoring needs
a proprietary environment (the free Engine is licensed for development and
non-commercial use), unlike Csound, MusicXML, and OSC. The dependency ends at
authoring: emitted \texttt{.csd}/\texttt{.orc}, MusicXML, and JSON are open
plain text. A language-neutral store serialization and Python/Julia core are
the intended portability path; until the port and paclet release exist, this is
future work, not present accessibility.

\section{Anatomy of the Paclet}

The implementation comprises fourteen modules exposing 80 public functions in
three \emph{module groups}: \texttt{core} (entities, store, tempo, session),
\texttt{rendering} (dispatcher and four backends), and \texttt{util} (pitch,
rhythm, dynamics, validation, graphics). Its separate conceptual taxonomy has
three \emph{architectural layers}: temporal, semantic, and rendering-contract.
Utilities support the layers rather than forming a fourth one. The first two
layers are backend-agnostic; adding a backend adds a Layer-III contract. All
functions are pure; store operations return new stores.

\textbf{Layer I.} Tempo is an ordered list of $\{$beat, BPM$\}$ points; each
BPM applies stepwise until the next point. Compilation integrates these
constant segments into a piecewise-linear $\{$beat, seconds$\}$ map with
analytical inversion. Continuous accelerando/ritardando is not supported.
Onsets and durations remain Rationals until a renderer requests seconds or
samples.

\textbf{Layer II.} The store groups entities by type
(\texttt{stemData[type][id]}, $O(1)$ typed access); its per-event rendering
pass is $O(n)$, but no large-score benchmark is claimed from the present
fixture. Seven implemented
entity types (note, rest, trigger, curve, state, marker, annotation), with four
more (process, interval, trajectory, gesture) declared for future use. A note carries pitch (spelling plus cents, MIDI float,
or frequency), duration, dynamic, and a \texttt{rendering} association keyed by
backend id---the one place backend-specific payloads travel with the entity.
Meter is a state entity: measure structure is authored, not guessed. A session
configuration carries everything a renderer may not hardcode, starting with
\texttt{tuningA4} (default 440~Hz, freely reassignable).

\textbf{Layer III.} Constructors validate only the shared shape and require each
\texttt{rendering[backend]} payload to be an association whose keys belong to the
backend contract; malformed Csound payloads are therefore local Csound errors.
Each backend declares a fallback policy: for example, MusicXML warns when it
reduces a continuous curve, which has no direct notation equivalent, to a
textual direction rather than silently inventing notation.

\begin{figure}
\centering
\includegraphics[width=0.62\linewidth]{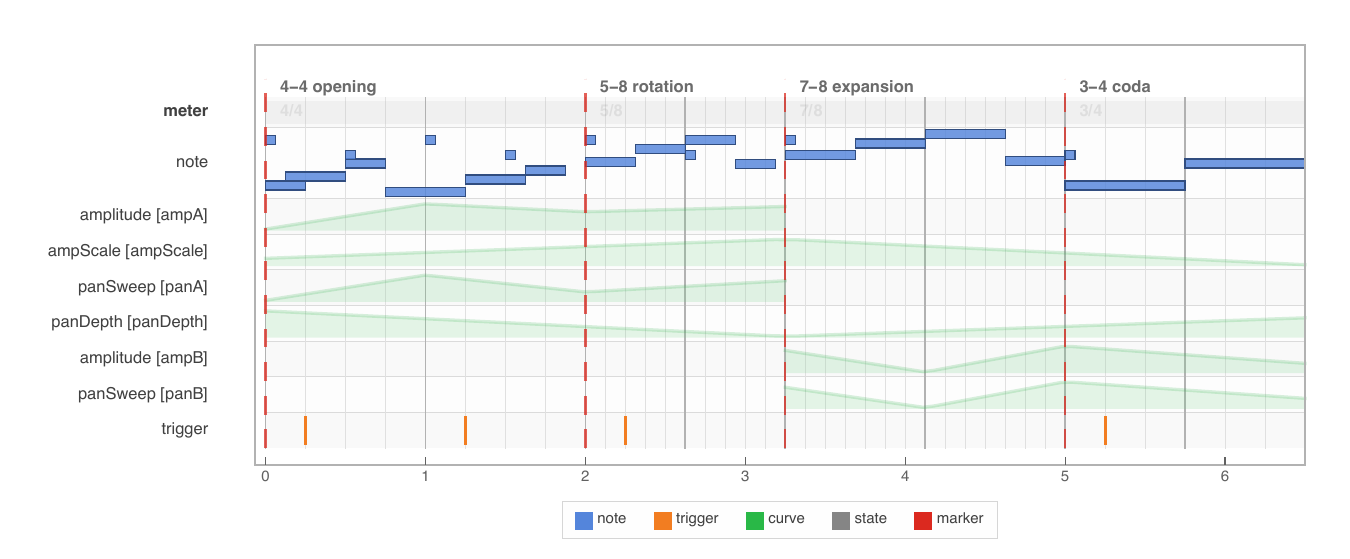}
\caption{A timeline exported directly by the paclet
(\texttt{drawTimeline}): a survey of the renderer's main
objects---sixteen \texttt{sine\_osc} notes (several with cent offsets),
eight \texttt{click} hits, four triggers, six overlapping amplitude/pan
curves, section markers, and a $4/4 \to 5/8 \to 7/8 \to 3/4$ meter grid. A
stress fixture exercising many entity types at once, not a representative
composition.}
\label{fig:timeline}
\end{figure}

The paclet also inspects itself: \texttt{drawTimeline} renders any stemData as
lane-based graphics with the metric grid derived from the authored meter
entities (Fig.~\ref{fig:timeline}), from a single function call with no external
tool. Every fixture in the Csound suite compiles without error and was
auditioned. Pending release of the paclet, the Zenodo
supplement~\cite{supplementary} covers most of the present reproducibility gap:
it archives the generating notebook, Csound/OSC/MusicXML/click outputs, audio,
and the tempo-edit synchrony demonstration used in
Sects.~\ref{sec:csound}--\ref{sec:sync}.

\section{The Implemented Renderers}

A backend is data: an association with an id, a renderer, a serializer, the
primitive types it emits, and a fallback policy. The generic dispatcher
\texttt{renderTo[stemData, tempo, backend, config, registry]} is the single
entry point. Four backends exist. \emph{Csound} compiles the whole stemData to
score and orchestra primitives (Sect.~\ref{sec:csound}); \emph{Click} derives a
standalone rehearsal track from meter and tempo alone (Sect.~\ref{sec:click}).
\emph{MusicXML} builds measures from meter state, splits notes at barlines into
tied fragments, and validates measure capacity, part coverage, and tie and
chord integrity; it is an explicitly lossy projection with a declared coverage
profile \cite{good}, currently in beta, emitting MusicXML~4.0 (partwise) and
checked by round-tripping through Dorico for tie, meter and part-separation
fidelity---the open, standards-conformant format a growing symbolic-music
ecosystem both produces and consumes~\cite{omr,bmdataset,text2score}.
\emph{OSC} emits a
schedule-ready JSON event list in milliseconds---curves sampled at a
configurable control rate---for dispatch as OSC messages \cite{osc} to whatever
is listening, a graphical patcher or a running Csound instance through its OSC
opcodes. Not every entity exists everywhere: curves become k-rate control
signals in Csound but reduce to a textual direction in MusicXML. MusicXML also
requires an authored spelling (or spelling plus cents) and rejects a bare MIDI
float or frequency. For integer MIDI, the explicit \texttt{midiToSpelling}
utility uses a documented sharp spelling (e.g., 58 becomes A\#3); the author
must invoke it, so a C\#/D$\flat$ choice is never hidden. The diatonic synchrony
fixture does not test this policy. Such differences are recorded per
entity--backend pair rather than treated as lossless.

\section{The Csound Renderer}\label{sec:csound}

\begin{figure}
\centering
\resizebox{\linewidth}{!}{\begin{tikzpicture}[
  font=\footnotesize\linespread{0.8}\selectfont,
  box/.style={draw, align=center, inner sep=2pt, fill=white},
  data/.style={box, fill=gray!8},
  arrow/.style={-Latex}
]
\node[box, text width=28mm] (rend)
  {\texttt{renderCSnd}\\[-3pt]
   {\scriptsize note/trig $\to$ \texttt{i}; marker $\to$ comment;
    curve $\to$ k-signal on p-fields}};
\node[box, text width=28mm, below=6mm of rend] (click)
  {\texttt{renderClickTrack}\\[-3pt]
   {\scriptsize meter grid $\to$ downbeat/beat clicks}};
\node[data, text width=21mm, minimum height=27mm,
      left=6mm of $(rend.west)!0.5!(click.west)$] (store)
  {entity store $+$ tempo map\\[-3pt]
   {\scriptsize rational beats; session config; orchestra registry}};
\node[data, text width=27mm, right=6mm of rend] (prim)
  {score $+$ orchestra primitives\\[-3pt]
   {\scriptsize p1 name\; p2 s\; p3 s\; p4 Hz\; p5 amp\; p6+;
    \texttt{f} tables, comments, k-globals, bound \texttt{instr}s}};
\node[data, text width=27mm, right=6mm of click] (cprim)
  {click events\\[-3pt]
   {\scriptsize strong 1760 Hz / weak 1100 Hz}};
\node[box, text width=17mm, right=11mm of $(prim.east)!0.5!(cprim.east)$] (ser)
  {\texttt{serialize-\\CSnd}};
\node[data, text width=27mm, right=6mm of ser] (csd)
  {\texttt{.csd}: score $+$ generated code $+$ \texttt{\#include}d
   \texttt{.orc}};
\node[box, dashed, text width=27mm, below=5mm of csd] (wav)
  {{\scriptsize optional \texttt{csound -W -o}: \texttt{.wav} at session rate}};
\draw[arrow] (store.east|-rend.west) -- (rend.west);
\draw[arrow] (store.east|-click.west) -- (click.west);
\draw[arrow] (rend.east) -- (prim.west);
\draw[arrow] (click.east) -- (cprim.west);
\coordinate (bus) at ([xshift=-4mm]ser.west);
\draw[arrow] (prim.east) -- (bus|-prim.east) |- ([yshift=2mm]ser.west);
\draw[arrow] (cprim.east) -- (bus|-cprim.east) |- ([yshift=-2mm]ser.west);
\draw[arrow] (ser.east) -- (csd.west);
\draw[arrow] (csd.south) -- (wav.north);
\end{tikzpicture}}
\caption{Both Csound-family renderers share one serializer; source
instruments stay in external \texttt{.orc} files, and curve bindings add
generated copies without touching them.}
\label{fig:pipeline}
\end{figure}
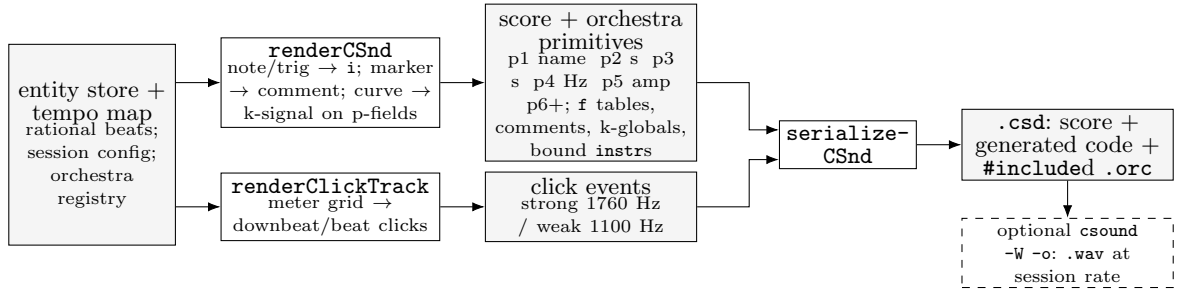

Csound is the oldest backend and fixes the paclet's rendering idiom, carrying
the electroacoustic and synthesis output (Fig.~\ref{fig:pipeline}). Its default
p-field contract is \texttt{p1} instrument, \texttt{p2} onset seconds,
\texttt{p3} duration seconds, \texttt{p4} frequency in Hz, \texttt{p5} amplitude
in $[0,1]$, and \texttt{p6}+ from the entity payload; session instrument
contracts can remap amplitude and provide intervening p-field defaults.
The renderer (\texttt{renderCSnd}) takes the whole stemData: notes and triggers
become \texttt{i}~statements, markers onset-ordered score comments, and each
curve a k-rate control signal in a global variable---an \texttt{f}-table reader
(a GEN-2 table read at k-rate by \texttt{tablei}) or a generated \texttt{transeg},
per session choice.

Two decisions matter most. First, \emph{late conversion}: because onsets
stay Rational until this point, \texttt{p2}/\texttt{p3} are computed from
the compiled tempo map, \texttt{p4} from spelling and cents under the
session's \texttt{tuningA4}, and \texttt{p5} from the dynamic marking, all
at render time. Retuning a session to 432~Hz or editing its tempo specification
regenerates every second and every frequency without touching one entity.
Second, \emph{named instruments}: an orchestra registry is loaded from a
directory of \texttt{.orc} files, the filename being the stable id used
both in \texttt{instr} definitions and in entity payloads. No numeric
renumbering ever occurs, so the note

{\small\begin{verbatim}
i "sine_osc" 0. 4. 440. 0.65 0.5
\end{verbatim}}

\noindent
(A4, \emph{mf}, two whole notes at 120~BPM, actual paclet output) remains
readable and stable as the orchestra grows.

Curves reach sounding notes through session-declared bindings. Each parameter
names a target p-field and a mode: \texttt{multiply} preserves and scales the
note's base value (the default amplitude binding), whereas \texttt{replace}
discards that base value (the pan sweep); later multiplicative bindings may
scale the replacement. For each affected instrument
the renderer copies the whole \texttt{instr} body from its source \texttt{.orc}
into a \emph{bound copy}, rewriting only the use-sites of the bound
p-fields---the declared control-signal expression replaces their use-sites, everything else
(\texttt{pcount()} guards, envelopes, \texttt{limit()} clamps, the oscillator)
reproduced verbatim. Affected notes are re-pointed to this copy, and several
curves may compose on one p-field (in Fig.~\ref{fig:timeline}, two amplitude
curves and an \texttt{ampScale} curve into \texttt{p5}; \texttt{panSweep} and
\texttt{panDepth} into \texttt{p6}). Regenerated every render and never edited
at the source, the copy cannot diverge through manual editing. The serializer writes the globals and
generated instruments beside the registry's \texttt{\#include} lines, sorts
score statements by onset, and can compile the \texttt{.csd} to audio. Csound
named software channels are a robust alternative when an orchestra explicitly
opts into channel names and voice-addressing conventions. The current p-field
adapter instead targets deterministic offline, self-contained score artifacts
and legacy score-style instruments; duplicating and rewrite-parsing an
instrument body is consequently a limitation, and a channel-based adapter is
future work.

The same amplitude curve lives differently outside Csound. Projected to OSC it
is not bound into an instrument but sampled at the session control rate into
timed messages. A two-second linear ramp sampled half-open at $10$~Hz contains
events from 0 through 1900~ms: its JSON \texttt{duration\_ms} is therefore 1900,
not 2000, and the last message has \texttt{time\_ms}~1900,
\texttt{address} \texttt{/param/amplitude}, \texttt{args}~\texttt{[0.95]}
(actual paclet output, excerpted from the supplement). The beat-space curve and tempo
map are the same as in the Csound projection; one binds the curve to a score
p-field, the other streams it as OSC messages in milliseconds---a genuinely
non-Csound artifact, the heterogeneity the architecture exists to make cheap.

\section{A Click Track from the Same Timeline}\label{sec:click}

Click tracks for mixed music are commonly rebuilt by hand in a DAW whenever the
score changes---precisely the divergence this architecture removes. The click
backend treats rehearsal audio as one more projection: it reads only the
authored meter entities and the tempo map, builds the measure grid, and emits
one short \texttt{i}~event per beat, accenting each downbeat. Every parameter is
session configuration---beat unit ($1/4$), click duration ($1/16$), strong/weak
frequencies ($1760/1100$~Hz), amplitudes ($0.9/0.45$), and pan---and the click
instrument itself is an ordinary three-line \texttt{.orc} file. Because the
renderer walks the same meter grid and tempo map as every other backend, meter
and tempo changes are followed with no additional authoring; for two $4/4$ bars
at 120~BPM it emits strong \texttt{i "click"} downbeats (0 and 2~s) and weak
intervening beats (full \texttt{.csd} and \texttt{.wav} in the supplementary
folder). In roughly 140 lines that reuse \texttt{serializeCSnd} unchanged
(Fig.~\ref{fig:pipeline}) and touch no line of Layers~I--II, the backend
inherits \texttt{\#include} handling and optional \texttt{.wav}
compilation. This simplest backend demonstrates reuse in one case; it is not a
general measure of the cost of adding arbitrary projections. The genuinely different
time encodings, notation's divisions and OSC's milliseconds, belong to the
MusicXML and OSC contracts.

\section{Keeping the Projections Synchronous}\label{sec:sync}

Because every backend reads the same rational onsets and the same compiled
tempo map, the projections cannot drift apart. Take one authored timeline---two
$4/4$ bars then $3/4$, $120$ then $90$~BPM---projected to Csound, click, OSC,
and the beta MusicXML backend. The
downbeat of bar~3 is Csound \texttt{p2}$=4.0$\,s, OSC \texttt{time\_ms}$=4000$,
and a strong click at $4.0$\,s, while MusicXML places it at bar~3, beat~1.
Editing the second tempo segment to $60$~BPM moves every later onset by the same
amount in all three time-domain backends (the note at beat $9/4$: $4.667 \to
5.0$\,s), while in the MusicXML export only the second tempo mark changes ($90
\to 60$~BPM)---the initial $120$ and every written note-duration unchanged:
notation lives in beat space, and seconds are bound only at render time.

\section{Conclusions}

Temporal System compiles an immutable rational-beat store into unequal but
synchronized Csound, click, notation, and OSC projections. The paclet is not
yet released and MusicXML remains beta. The system has no MIDI or continuous-
tempo support: MIDI is not currently emitted, but is planned for a future
implementation, alongside other renderers that may emerge as useful targets,
including ones not yet anticipated by the present design. Csound curves assume
declared p-fields and rewritten instrument copies; named channels are not
implemented. The unreleased Wolfram paclet also keeps authoring proprietary
although its outputs and supplement are portable. These boundaries delimit the
results and the open-source portability work above. The enduring separation of
computed time from rendered sound lets one timeline become many renderings.

\end{document}